\documentclass[conference]{IEEEtran}
\IEEEoverridecommandlockouts

\usepackage{multirow}
\usepackage{cite}
\usepackage{caption}
\usepackage{amsmath,amssymb,amsfonts}
\usepackage{url}
\usepackage{graphicx}
\usepackage{textcomp}
\usepackage{xcolor}
\usepackage{amsmath}
\usepackage{algpseudocode}
\usepackage{algorithm}
\usepackage{subfig}
\usepackage[english]{babel}
\usepackage[utf8]{inputenc}
\usepackage{xcolor, soul}
\sethlcolor{green}
\usepackage{pdfpages}
\usepackage{booktabs}
\usepackage{graphicx}

\usepackage[nottoc]{tocbibind}
\algnewcommand\algorithmicforeach{\textbf{for each}}
\algdef{S}[FOR]{ForEach}[1]{\algorithmicforeach\ #1\ \algorithmicdo}
\usepackage{algorithm}
\usepackage{eso-pic}
\usepackage[acronym]{glossaries}
\newacronym{ntc}{NTC}{Network Traffic Classifier}
\newacronym{qos}{QOS}{Quality of Service}
\newacronym{crnn}{CRNN}{Convolutional Recurrent Neural Network}
\newacronym{lstm}{LSTM}{Long Short-Term Memory}
\newacronym{rnn}{RNN}{Recurrent Neural Network}
\newacronym{kde}{KDE}{Kernel Density Estimation}
\newacronym{nlp}{NLP}{Natural Language Processing}
\newacronym{cnn}{CNN}{Convolutional Neural Network}
\author{
    \IEEEauthorblockN{Matin Shokri \IEEEauthorrefmark{1}, Ramin Hasibi\IEEEauthorrefmark{2}}
    \IEEEauthorblockA{\IEEEauthorrefmark{1}
    Department  of  Electrical  and Computer Engineering\\
    K. N. Toosi University of Technology
    \\shokri@email.kntu.ac.ir}
    \IEEEauthorblockA{\IEEEauthorrefmark{2}Department of Informatics, University of Bergen, Bergen, Norway \\
    ramin.hasibi@uib.no
    }
}
\begin{document}

\title{
A Pipeline of Augmentation and Sequence Embedding for Classification of Imbalanced Network Traffic}
\maketitle

\begin{abstract}
Network Traffic Classification (NTC) is one of the most important tasks in network management. The imbalanced nature of classes on the internet presents a critical challenge in classification tasks. For example, some classes of applications are much more prevalent than others, such as HTTP. As a result, machine learning classification models do not perform well on those classes with fewer data. To address this problem, we propose a pipeline to balance the dataset and classify it using a robust and accurate embedding technique. First, we generate artificial data using Long Short-Term Memory (LSTM) networks and Kernel Density Estimation (KDE). Next, we propose replacing one-hot encoding for categorical features with a novel embedding framework based on the "Flow as a Sentence" perspective, which we name FS-Embedding. This framework treats the source and destination ports, along with the packet’s direction, as one word in a flow, then trains an embedding vector space based on these new features through the learning classification task. Finally, we compare our pipeline with the training of a Convolutional Recurrent Neural Network (CRNN) and Transformers, both with imbalanced and sampled datasets, as well as with the one-hot encoding approach. We demonstrate that the proposed augmentation pipeline, combined with FS-Embedding, increases convergence speed and leads to a significant reduction in the number of model parameters, all while maintaining the same performance in terms of accuracy.
\end{abstract}

\vspace{-2mm}

\section{Introduction}
\label{I. Introduction}
With the ever-increasing amount of traffic that flows through the Internet, network management has become a more demanding task. One of the most important aspects of network management is identifying the types of traffic in the network. Network traffic classifiers (NTCs) are applications responsible for detecting anomalies in the network and classifying network applications for Quality of Service (QoS) purposes \cite{maghalebasecrnn, deepbeliefintrusion}. These applications usually fall into one of the following categories based on their methodology, with each method having its benefits and drawbacks:

\begin{itemize} \item \textbf{Port-based:} These classifiers assign a specific port number to each application. However, some applications do not use a specific port (e.g., BitTorrent), and if the port is changed, this method is no longer reliable \cite{maghalebasecrnn}. \item \textbf{Deep Packet Inspection (DPI):} These applications classify traffic by inspecting the packet payload data through a rule-based approach. Two major drawbacks are that they need to be updated with new patterns in the payloads of applications when those applications modify their protocols \cite{maghalebasecrnn}. In addition, if they do not have access to the payload of packets for privacy reasons, their performance methods mentioned above are limited due to the lack of information about certain rules. \item \textbf{Machine learning-based:} The flaws of the two methods mentioned above have resulted in the growing popularity of the third type of classifier, which uses machine learning, specifically deep learning algorithms. These algorithms work with the features in the header of the packets, but some may also consider the information in the payloads \cite{maghalebasecrnn, sharifia}. Although they are still limited, they have shown great potential in terms of performance and could potentially be a great substitute for the aforementioned methods. \end{itemize}

Most of the traces gathered from real Internet traffic are imbalanced, meaning some types of application traffic are generally more prevalent than others (e.g., HTTP) \cite{elsevierimbalancedgravity, acganaug, devide&conqueraug, Shafiq2018}. This issue becomes more pronounced in large-scale traffic and can cause significant problems for algorithm performance on the sparse classes. An effective way to address this imbalance is through \textit{augmentation}. Augmentation is an approach in machine learning that tackles the issue of having too little data for training. It typically aims to increase the training data in a way that does not alter the nature of the sparse classes. Augmentation methods should be designed on the basis of the characteristics of the data. For instance, augmentation techniques used for image classification may not be effective in the network traffic domain.

State-of-the-art GAN-based methods have recently demonstrated strong performance on specific data types, such as images. However, in certain domains like network traffic, GANs may suffer from overfitting or noise, especially when working with numerical or categorical features with a range of discrete values including (e.g., port numbers) \cite{zhao2021ctab}. Furthermore, learning from these features often requires transforming them into other spaces more suitable for machine learning methods. One-hot encoding is a common technique for converting numerical and categorical data into formats suitable for machine and deep learning algorithms, and it generally improves predictions and classification accuracy. While one-hot encoding improves performance in specific contexts, It does not account for the semantic meaning of sequential data. For example, the order of ports and packet direction have sequential relationships in our application, which a simple one-hot encoding cannot capture.

To address these challenges, we introduce a novel pipeline that not only improves the accuracy and convergence speed of classification models but also reduces the number of model parameters required in deep learning algorithms trained on real-world traffic traces.

The remainder of this paper is organized as follows: In Section 2, we review the previous work in the area of \gls{ntc}. In Section 3, we describe our augmentation and embedding pipeline. The dataset and deep learning models used to classify the traffic traces are discussed in Sections 4 and 5, respectively. Finally, the evaluation of our method is presented in Section 6.\section{Related Work}

There are two approaches when it comes to classifying network traffic using machine learning. One approach considers each packet individually and makes predictions based on the features extracted from the packet itself. The other approach classifies a flow of packets (a stream of packets transferred between a pair of source and destination for a specific application), which has the benefit of capturing the sequential dependencies that arise from features extracted from the flow. This approach can also be processed using methods inspired by the field of Natural Language Processing (NLP). With the advent of deep learning as one of the most effective machine learning methods, achieving high accuracy in many fields, many recent studies have employed different neural network architectures to classify network traffic datasets. In \cite{maghalebasecrnn}, Lopez-Martin et al. present a deep Convolutional Recurrent Neural Network (CRNN) architecture to classify network flows and tune the hyperparameters of the model to find the optimal settings and feature set. However, they have not addressed the imbalance problem in their dataset. Additionally, the scale of the dataset used in their study does not reflect the real-world scale of internet traffic in terms of volume and the number of classes. In \cite{springer_deep}, Rahul et al. also proposed using Convolutional Neural Networks (CNN) to classify network traffic, but their work only considers three classes of applications on a limited dataset. In \cite{sharifia}, a comparison between CNN and Stacked Auto-Encoders for classifying both types of traffic and applications in the network in a standard VPN/none-VPN dataset is presented. In this method, instead of flows, each packet is classified individually based on the features from both header and payload of packets which may not be available in some privacy-preserving datasets. In \cite{bayesianneural}, Auld et al. deploy a Bayesian neural network in the form of a multi-layer perception and classify their dataset accordingly. In this work, the proposed method tends to underperform when it comes to classes with the lowest number of data in the dataset. 
Transformers \cite{vaswani2017attention} are state-of-the-art deep learning models that use the self-attention mechanism to weigh the relevance of each input data element differently. They are primarily used in natural language processing (NLP) and computer vision. Recently, Transformers have also been widely used in Network Traffic Classification (NTC) tasks. PERT \cite{he2020pert} is a Transformer-based model that uses the payload of packets to classify encrypted HTTPS traffic data. The Multi-task Transformers (MTT) \cite{zheng2022mtt} is a multi-task learning approach that simultaneously classifies traffic characterization and application identification tasks. To extract features, the proposed model treats the input packet as a sequence of bytes and employs a multi-head attention mechanism. FlowFormers \cite{FlowFormers} enhances the FlowPrint representation with attention-based Transformer encoders. This model outperforms traditional deep learning models on NTC tasks such as application type and provider classification.

Some previous work has attempted to address the imbalance issue in network traffic datasets. In \cite{PGMMOORE}, Rotsos et al. introduced a method using Probabilistic Graphical Models for semi-supervised learning in a Naive Bayes model. For their learning, they assumed a Dirichlet distribution before the classes with a high $\alpha$ value. This assumption is based on the idea that some classes have a higher probability than others. In \cite{acganaug}, an augmentation method is proposed using an Auxiliary Classifier Generative Adversarial Network (AC-GAN). However, only two classes of network traffic are considered: SSH and non-SSH, and the dataset used does not extend to other network applications. Furthermore, the method is only evaluated on traditional machine learning algorithms such as Support Vector Machines, Random Forest, and Naive Bayes. Additionally, \cite{devide&conqueraug} presents a new feature extraction method using a divide-and-conquer approach for handling imbalanced datasets in network traffic.

For augmentation purposes, one needs to simulate the sequential patterns of flows through time series generation methods. LSTMs \cite{LSTM}, as one of the most popular methods in sequence representation learning, have also been utilized for generating sequential data. For example, \cite{LSTMGEN} introduced a method to generate data using LSTMs and evaluated the method to show that it can capture the temporal features in the dataset. LSTMs have also been used as an augmentation tool in works such as \cite{daskhataug} and \cite{LSTM_ae_aug} for generating handwriting and human movement data, respectively, and have proven to be effective in both cases.

Embedding methods for sequences of words are critical in transforming raw text data into numerical representations that machine learning models can effectively process. These embeddings, such as Word2Vec, GloVe, or transformer-based models like BERT, capture semantic relationships and contextual meaning between words, allowing models to understand the underlying structure of language. By converting words into dense vectors in a continuous vector space, embeddings reduce the dimensionality of the input data, leading to more efficient training and faster convergence. This transformation helps models learn more effectively, as they can focus on meaningful features instead of raw word counts or one-hot encodings. As a result, embedding methods not only speed up the convergence process but also improve the overall performance of models, enabling them to generalize better on unseen data and handle complex tasks with higher accuracy.

Flows in a network resemble sentences in a language. As a result, word embedding techniques can be useful when representing a network flow. In word embedding \cite{turian2010word}, words with similar meanings are represented similarly in a learned representation of text. One advantage of adopting dense, low-dimensional vectors is the reduction of computational costs. Most neural networks do not operate well with very high-dimensional, sparse vectors, such as one-hot encoding, especially when the number of categories is enormous \cite{li2018slim}. Hence, a neural network must perform computationally expensive operations to properly learn a representation that relates each word to another. An embedding layer \cite{gal2016theoretically} is a word embedding approach that is learned alongside a neural network model. The embedding layer allows us to convert each word into a fixed-length vector of a specific size. The resulting vector is dense, with real values, rather than just 0’s and 1’s. The fixed length of word vectors allows us to better represent words while reducing dimensionality. To the best of our knowledge, previous work in this area employs models that are either too computationally expensive or fail to consider performance on the sparse classes in the dataset. We aim to address these issues through our proposed augmentation scheme, which uses LSTM to generate traffic flow patterns that balance the dataset, as well as a flow packet embedding scheme that enhances both the speed and accuracy of the deep learning model.

\section{Methodology}
In this section, we introduce our proposed method for augmenting network traffic flow datasets, the embedding scheme for traffic flow packets, and the Transformer architecture used to predict traffic flows.

\subsection{Augmentation Scheme for Generating Time Series Network Data}
In this section, we describe our augmentation scheme for generating new data from network traffic traces.

Every flow in the network has the same 5-tuple attributes: \begin{itemize} \item Source and destination IP addresses. \item Source and destination port numbers. \item Link layer protocol (e.g., TCP, UDP). \end{itemize} Every application on the internet creates a flow of packets between communicating peers \cite{naivebayesaggregate}.

To represent flows in our work, we select a set of features for each flow that can capture its nature. These features can be sequential or numerical and are proposed as a sample set of features for a machine learning-based \gls{ntc}. These features can also be extended to further enrich the input information for each flow. However, our goal is to demonstrate how to augment each category of features, not to determine the optimal set of features for classification. The set of features used for classifying flows in our work are listed in Table 1. These features are gathered from the first 20 packets of each flow, serving as a cutoff point for the flow.

\begin{table}[t!]
\caption{Features of each flow}
\label{table:featuretable}
\begin{center}
\begin{tabular}{ |p{2.5cm}||p{1.25cm}| } 
 \hline
 \textbf{Feature} & \textbf{Type} \\
 \hline
Source port & \multirow{4}{4em}{Numerical} \\ 
Destination port & \\
Inter-arrival time &\\
Payload length &\\
 \hline
Direction of packet & \multirow{2}{4em}{Sequential} \\ 
TCP window size & \\
\hline
\end{tabular}
\end{center}
\end{table}
Each group of features (numerical or sequential) has its way of augmentation which is described in the following.
\subsubsection{Generating sequential features}

As mentioned earlier, a traffic flow consists of the sequence of packets transmitted between a source and a destination. Some applications are uni-directional, i.e., the packets are transmitted in only one direction (e.g., uploading data). However, in some applications, packets flow in both directions, such as when a client communicates with a server and receives a response to its request. Whether a packet is sent from the source or the destination depends on the sequence of packets that have already been transmitted in the flow. Therefore, we can conclude that the sequence of packet directions in a specific application follows a time-series pattern and can be generated using sequence generation techniques, as seen in \cite{daskhataug, LSTM_ae_aug}. TCP window size is another feature of the flow that depends on the previous values in the flow. Generally, this value serves as an indicator of the connection's condition and the processing speed of data in the flow \cite{rfc1323}. Thus, its value at each step in the flow is affected by the values of previous steps. The sequence pattern generation scheme requires training an LSTM network to generate instances for each class separately. Each LSTM block attempts to learn the probability distribution at each step of the sequence while considering information from previous steps. To train the network, we gathered the patterns of packet directions in a flow for up to 20 packets within each class of flow applications. We encode each direction as 1 or 0, where 1 represents a packet sent from source to destination and 0 represents the opposite direction. At the end of each sequence, we add a unique character to indicate the end of the flow. Then, each sequence is shifted by one character to the right and used as the label to train the prediction model for each step. During the generation phase, we first choose a direction based on the distribution of that direction in the dataset for the first time step. This value is input to the LSTM, which then samples from the output probability distribution at each step (which could be 1, 0, or the ending character) to generate a new step. The generated direction is then fed into the LSTM to generate the probabilities for the next step. The maximum number of steps is 19, generating a flow pattern of up to 20 packets (the first packet is always from source to destination). For generating window size values, we use the same scheme, but in this case, the "characters" are the window size values from the dataset, rather than 0 and 1.

\begin{figure}[t!]
\centering
\includegraphics[width=7.70cm, height=6cm]{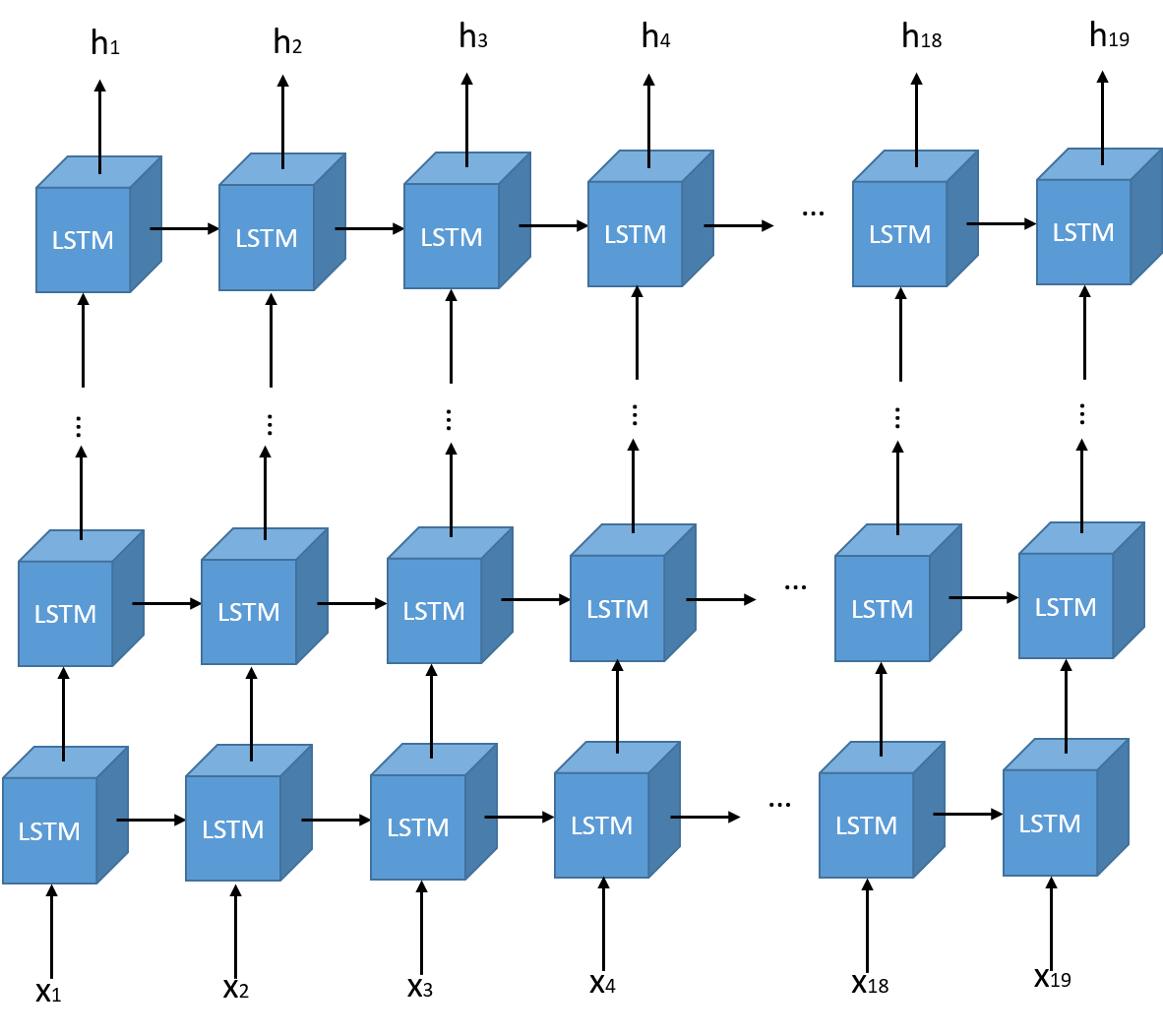}
\caption{Generating 20 time steps of packet sequence with LSTM}
\label{fig:generate_pakets}
\end{figure}
\subsubsection{Generating numerical features}

In this section, we describe our method of generating numerical features of a flow.

In this section, we describe our method for generating numerical features of a flow.

As shown in Table 1, there are four numerical features that we consider for each packet of the flow. To generate new samples from these features, we first need to estimate their probability distribution. Since these features are not sequential, we can use conventional probability density estimation methods. One such method is Kernel Density Estimation (KDE), which belongs to the category of kernel methods.

KDE, also known as the Parzen–Rosenblatt window, is one of the most well-known methods for estimating the probability density function (PDF) of a dataset. As a non-parametric density estimator, KDE makes no assumptions about the form of the density function, unlike parametric algorithms. This method automatically learns the shape of the density from the data. The flexibility of KDE, arising from its non-parametric nature, makes it particularly popular for data drawn from complex distributions.

Let $X = {x_1, x_2, ..., x_n}$ denote the set of independent, identically distributed random samples from a group of features (e.g., inter-arrival time), and let $K(x): \mathbb{R}^d \mapsto \mathbb{R}$ denote the probability distribution function (PDF) of the kernel of our choosing. We can estimate the PDF of $X$ by:

\begin{equation}
    \hat{p} = \frac{1}{nh^d} \sum_{i=1}^{n}K(\frac{x-x_i}{h})
\end{equation}

Where $\hat{p}$ is the estimated PDF of $X$ and $h>0$ is the bandwidth of the kernel that is used to control the smoothing degree of the kernel\cite{Rosenblatt,parzen1962}.

One common example of K(x) is Gaussian distribution with $\mu = 0$ and $\sigma = 1$ as expressed by
\begin{equation}
    K(x) = \frac{1}{{\sqrt {2\pi } }}e^{{{ - \left( {x  } \right)^2 } \mathord{\left/ {\vphantom {{ - \left( {x - \mu } \right)^2 } {2\sigma ^2 }}} \right. \kern-\nulldelimiterspace} {2}}}
\end{equation}
which is also used for our scheme.

To prevent bias-variance problems in fixed $h$ cases, we used the bandwidth selection method represented in \cite{silverman}. According to \cite{silverman} if a Gaussian kernel is used, it can be shown that the optimal value of $h$ is $h^*=\Big(\frac{4\hat{\sigma}^5}{3n} \Big)^\frac{1}{5}$.

\subsection{Embedding scheme for extracting feature from time series network data}

One-hot encoding is a standard procedure used to prepare categorical features for input into a machine-learning algorithm. However, as the number of possible values for categorical features increases, the dimensionality of the encoding—and, as a result, the required memory to store these features—grows significantly. Moreover, models trained on such embeddings are prone to the curse of dimensionality and sparsity. For example, port numbers in a network traffic application can range from 0 to 65535, meaning 65536 bits would be needed to create a one-hot feature. Word embedding, a technique commonly used for text analysis, encodes a word’s meaning in such a way that words with similar contexts are represented by vectors that are close together in the vector space \cite{bengio2000neural}. Based on this technique, we propose a scheme to represent specific features, such as ports and packet directions, as vectors suitable for neural network models. This allows the model to learn associations between ports across a large number of flows. In our scheme, we first map each port to a character based on its frequency within each application. This results in the creation of a lookup table that links each port to a corresponding character. In this paper, we consider 19 applications, and each port is mapped to one of these applications. For example, port number 80, which is most frequent in the HTTP class, is assigned to the "A" character. A part of this lookup table is shown in Table \ref{table:lookuptbl}.

\begin{table}[htbp]
\caption{Lookup Table Schema}
\label{table:lookuptbl}
\begin{center}
\begin{tabular}{ |c|c|c|c|c|c|c| }
\hline
\multicolumn{1}{|c|}{}&
\multicolumn{6}{|c|}{Character}\\

\hline
Port&A&B&C&...&R&S\\

\hline
\rotatebox[origin=c]{90}{...}&
\multicolumn{6}{|c|}{\rotatebox[origin=c]{90}{...}}\\
&
\multicolumn{6}{|c|}{}\\

\hline
80&*& & & ... & &\\

\hline

\rotatebox[origin=c]{90}{...}&
\multicolumn{6}{|c|}{\rotatebox[origin=c]{90}{...}}\\

&
\multicolumn{6}{|c|}{}\\

\hline

\end{tabular}
\end{center}
\end{table}

FS-Embedding is a word embedding scheme based on the "Flow as a Sentence" perspective, which assumes that each flow consists of meaningful subsequences that convey parts of the overall information in the full sequence. This scheme treats each subsequence as a "word" that should have a semantic relationship with other parts of the flow. For example, Table \ref{table:flowtable} illustrates an HTTP class that includes Source Port (SP), Destination Port (DP), and Packet Direction (PD). The sentence is formed by concatenating the replaced SP, DP, and PD from a flow. These operations are integrated into a layer called the "generalizer." Once the "words" are created, an embedding layer learns from them during the classification task.

\begin{table}[htbp]
\caption{Example of Generating Words from SP, DP, PD}
\label{table:flowtable}
\begin{center}
\begin{tabular}{ |c|c|c|c|c|c|c|c|c|c| }
\hline
\multicolumn{10}{|c|}{Flow}\\

\hline
\multicolumn{3}{|c|}{Subsequence 1} &\multicolumn{3}{|c|}{Subsequence 2}&\multicolumn{1}{|c|}{...}&\multicolumn{3}{|c|}{Subsequence N}\\

\hline
SP & DP & PD & SP & DP & PD & ... & SP & DP & PD\\

\hline
80 & 443 & 0 & 443 & 80 & 1 &... & 443 & 80 & 1\\

\hline
A & E & 0 & E & A & 1 &... & E & A & 1\\

\hline
\multicolumn{3}{|c|}{Word 1} &\multicolumn{3}{|c|}{Word 2}&\multicolumn{1}{|c|}{...}&\multicolumn{3}{|c|}{Word N}\\

\hline
\multicolumn{3}{|c|}{AE0} &\multicolumn{3}{|c|}{EA1}&\multicolumn{1}{|c|}{...}&\multicolumn{3}{|c|}{EA1}\\
\hline

\end{tabular}
\end{center}
\end{table}

The FS-Embedding process consists of the following steps:
\begin{itemize} \item Generate a lookup table based on the frequency of each port in each application. \item Replace ports with characters according to the lookup table. \item Concatenate the source port, destination port, and packet direction to generate a word. \item Represent the flow as a sentence. \item Train an embedding layer using the sentences during the classification task. \end{itemize}

Fig \ref{fig: FS-Embedding} demonstrates the embedding process. The generalizer layer replaces ports considering its lookup table. After that, this layer concatenates characters and packet direction to generate a word. Then these words are fed to the embedded layer to generate the embedded vectors. Finally, vectors and other features generate a new flow to provide an input data model.

\begin{figure}[h]
    \centering
    \includegraphics[scale=0.28]{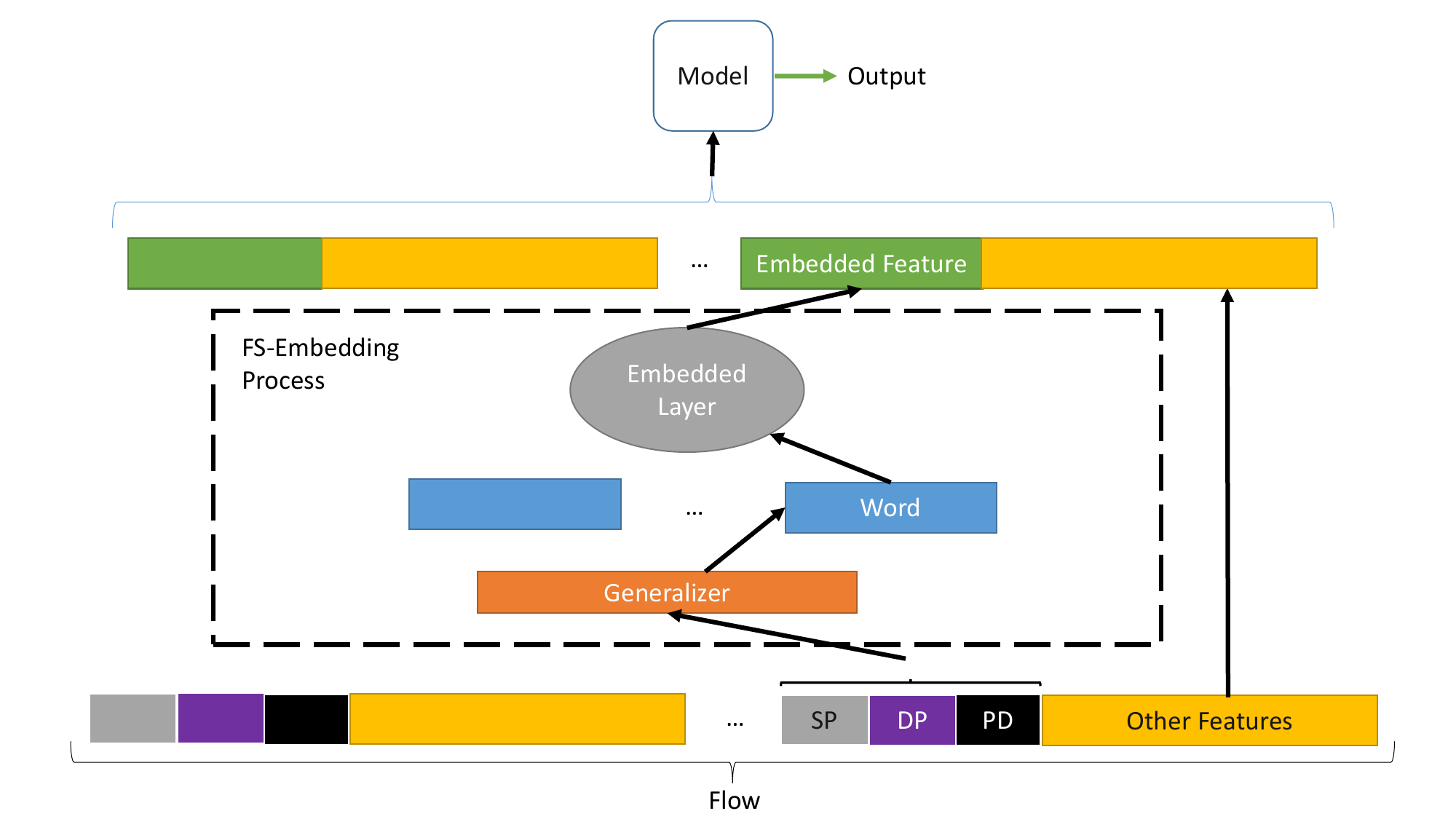}
    \caption{FS-Embedding Process}
\label{fig:FS-Embedding}
\end{figure}

\subsubsection{Transormer architecture}

Transformers are deep neural network models that utilize self-attention mechanisms to develop models for neural machine translation. The primary goal is to achieve high parallelization by removing the convolutional or recurrent blocks present in traditional neural machine translation models. As a result, transformers enable significantly more parallelization and can reach new levels of translation quality. The transformer design is based on a self-attention encoder-decoder structure, replacing recurrence and convolution layers for generating output. The attention module in the transformer performs computations in parallel multiple times, and the results are then combined to produce a final attention score. This approach, known as multi-head attention, allows the transformer to capture various relationships and complexities for each word.

The encoder maps an input sequence to a sequence of continuous representations, which is then fed into a decoder. The decoder receives the result of the encoder and its output at the previous time step to produce an output sequence. The model is auto-regressive at each phase, taking previously created symbols as additional input when creating the next. For shorter sequence lengths, self-attention layers are faster than recurrent layers, and for very long sequence lengths, they can be constrained to consider only a neighborhood in the input sequence.

\begin{figure}[h]
    \centering
    \includegraphics[scale=0.25]{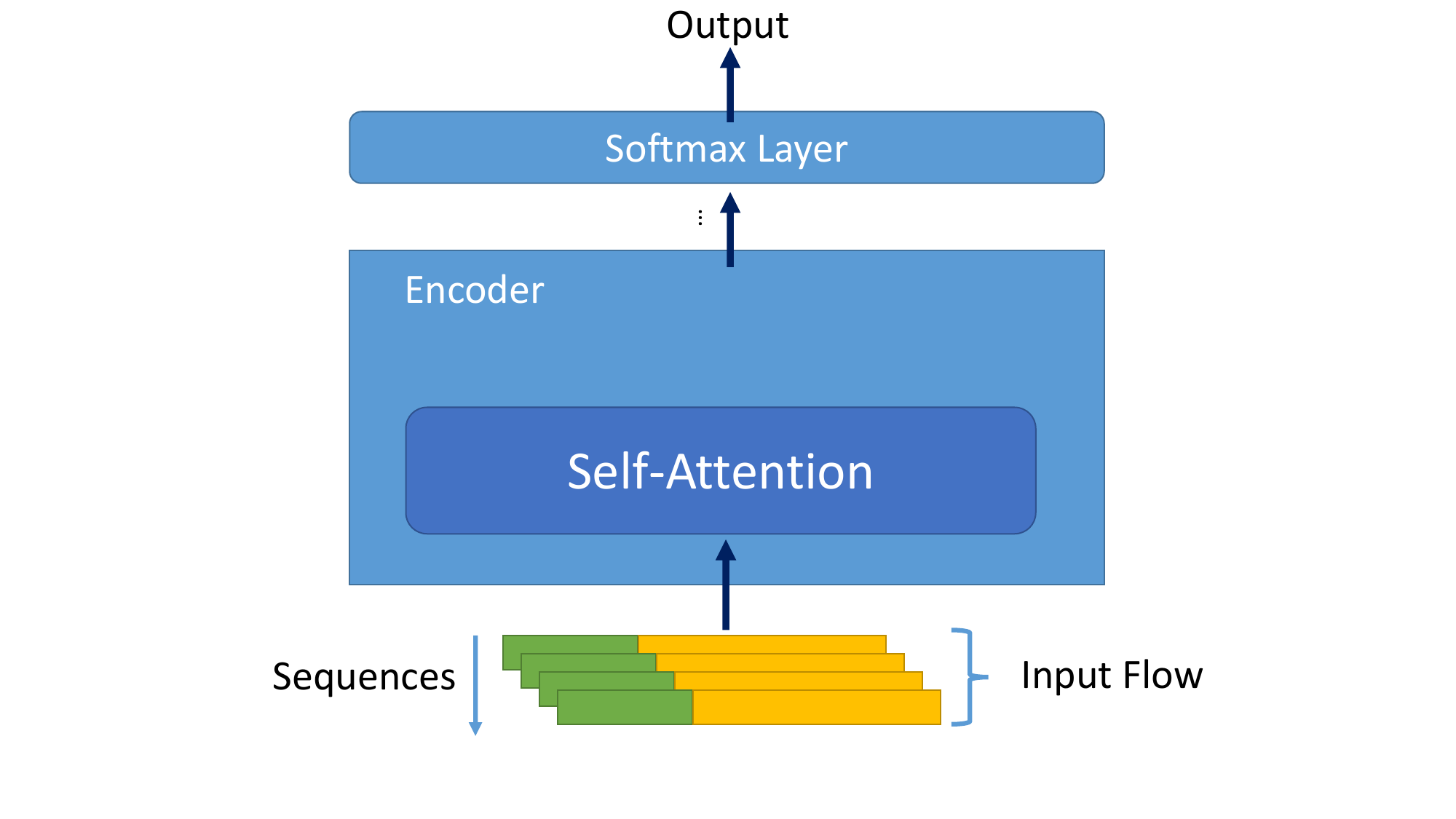}
    \caption{Classifier}
\label{fig:classifier}
\end{figure}

In this paper, we use the transformer's encoder part to develop our model as a classifier. The flow input, which consists of stacked sequences, is fed to the encoder input. Then the sequential features generated by the encoder will be classified by the softmax layer at the top of the encoder model. Fig \ref{fig:classifier} shows the mentioned process.

\section{dataset}
\label{datasetsec}
In this section, we describe the method used to gather and label our network trace dataset.

For this study, we utilized real network traffic traces from the campus of Amirkabir University of Technology, consisting of over 70 gigabytes of packets from both UDP and TCP link layer protocols. We labeled the flows using nDPI, an open-source Deep Packet Inspection (DPI) tool developed by ntop, which classifies flows based on applications \cite{nDPI}. The reason for choosing nDPI as our labeling tool is that, according to \cite{NDPICOMP}, nDPI is the most accurate open-source DPI tool available. We selected 19 traffic classes from over 50 gigabytes of packets, resulting in a total of 904,490 flows. Of these, 85 percent were allocated for training, while the remaining 15 percent were reserved for testing. The classes of applications used in the dataset correspond to those with the highest number of instances, and they are listed in Table \ref{table:classes}.

\begin{table}[htbp]
\caption{Classes of applications}
\label{table:classes}
\begin{center}
 \begin{tabular}{||c c c||} 
 \hline
\textbf{Label}&\textbf{Class} & \textbf{Number of Flows}\\
\hline
\hline
0 & HTTP & 58774 \\
1 & DNS & 126960 \\
2 & NTP & 4633 \\
3 & BitTorrent & 6146 \\
4 & HTTP\_Download & 16326 \\
5 & SSL\_No\_Cert & 10603 \\
6 & Steam & 4460 \\
7 & RDP & 1425 \\
8 & SSL & 341846 \\
9 & SSH & 9746 \\
10 & Facebook & 2772 \\
11 & Twitter & 2198 \\
12 & Google & 96072 \\
13 & WindowsUpdate & 2343 \\
14 & Telegram & 186256 \\
15 & Instagram & 6683 \\
16 & Microsoft & 18196 \\
17 & PlayStore & 5304 \\
18 & YouTube & 3747 \\
 \hline

\end{tabular}
\label{tab1:population}
\end{center}
\end{table}

As shown in Table \ref{tab1:population}, there are different classes of applications in our dataset and the names of our labels are chosen based on the labels given by nDPI.

\begin{figure}[htbp]
\centerline{\includegraphics[width=9.70cm, height=8cm]{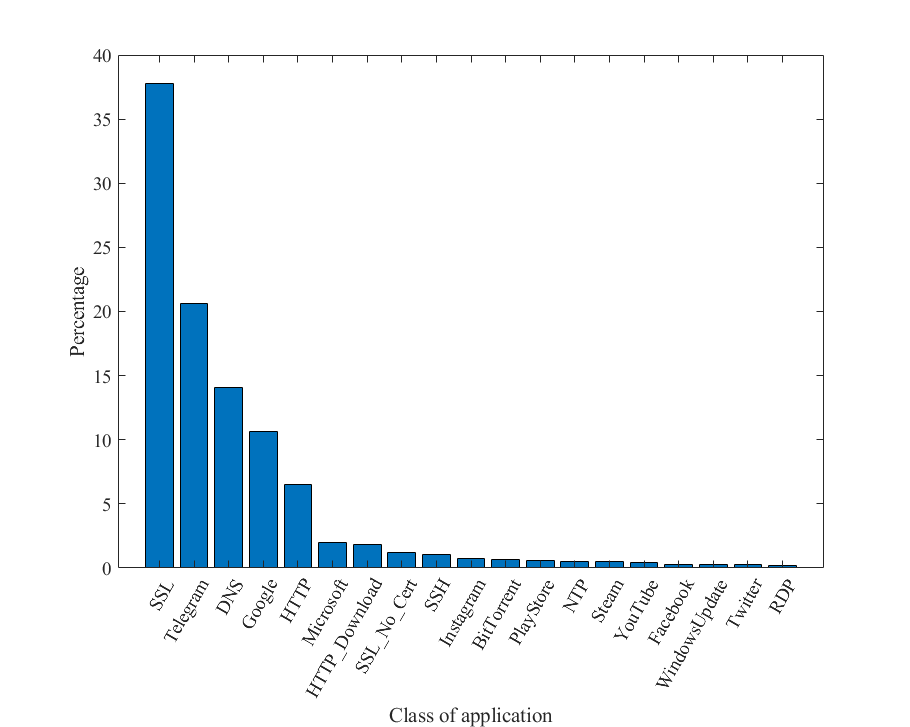}}
\caption{the percentage of different classes of applications in our dataset.}
\label{fig:populationpercent}
\end{figure}

The percentage of each class is shown in Fig. \ref{fig:populationpercent}.
As shown in the bar chart, the imbalance in the dataset is quite apparent. The most populated application class is SSL, which comprises more than 37\% of the total population, while the least populated class is RDP, accounting for less than 0.16\%. Additionally, over 86\% of the entire dataset is concentrated in just four classes. Moreover, eight classes represent less than 1\% of the data, making them the underrepresented classes that are likely to experience lower evaluation metrics.

\section{classification pipeline}

This section explains the classification pipeline used to test our augmentation and FS-Embedding performance.

The classification process consists mainly of two main stages:
\begin{itemize}
    \item Augmentation phase
    \item Training phase
\end{itemize}
\subsection{Augmentation} 
\label{Augmentation}
In the augmentation phase, we generate new data from classes with less population in the dataset. First, we train and use LSTM to create the pattern of directions and TCP window sizes in the flow. After that, we estimate the PDFs of every numerical feature using KDE. Then, according to these PDFs, we generate points in every feature domain. These points are our generated features for the packets. Finally, we generate up to 20 packets per flow and put these features in an array of size 6*20 (6 features from 20 packets). If the number of packets in the generated sequence is less than 20, the rest of the array is appended with 0. These arrays will comprise the generated dataset.

The pseudo-code for the augmentation process is given in Algorithm \ref{alg:the_alg}

\begin{algorithm}[t!] 
\caption{Augmentation process}
\label{alg:the_alg}
\hspace*{\algorithmicindent} \textbf{Input} set $C$ of classes with low population \\
\hspace*{\algorithmicindent} \textbf{Output} set $\hat{C}$ of generated flows
\begin{algorithmic}[1]
\ForEach {$c \in \mathcal C $}
\State $p \gets $ pattern of directions and TCP windows in $c$
\State Train LSTM for each pattern in $p$
\State $\hat{p} \gets $ generated patterns from trained LSTM
\State $NF \gets $ sets of Numerical features in $c$
\ForEach {set $nf \in  NF$}
\State  $PDF(nf)\approx KDE(nf)$
\State  $rs \gets $ generated random samples from $PDF(nf)$
\EndFor
\State $gen\_flows \gets$ new data from $rs$ and $\hat{p}$ based on \cite{maghalebasecrnn}
\State $\hat{C} \gets \hat{C} \cup gen\_flows$
\EndFor
\end{algorithmic}
\end{algorithm}

\subsection{Training}
In this section, we first compare the baseline model proposed in \cite{maghalebasecrnn} with the state-of-the-art Transformer model \cite{vaswani2017attention}, using both augmented data and the one-hot encoding scheme. Our best model consists of three encoder blocks, each featuring multi-head attention with four heads and an embedding size of 128.
The feedforward part of each encoder consists of 2 convolution layers of 4*1*1 and 19*1*1, respectively. Global average pooling performs on the output of the last encoder and then the result is fed into fully-connected layers with 512, 256, and, 19 hidden nodes. The baseline model (Convolutional Recurrent Neural Network) architecture includes two Convolution layers of 32*4*2 and 64*4*2, respectively. Each of these layers is followed by a batch normalization layer. At the end of the architecture, the output of the last batch norm is put in time-series form and is fed into an LSTM layer of 100 hidden units. There are two fully connected layers, 100 and 108 hidden nodes, and dropout rates of 0.2 and 0.4, respectively, followed by a soft-max layer with 19 outputs, each corresponding to 19 classes of traffic. For the second model, we use the encoder part of transformers for the classification task. The model consists of 3 stacked encoder blocks, ten heads with a size of 32. Then the encoder's output feeds to a fully connected neural network of 512 and 256 hidden neurons and 19 dense output under a soft-max layer.

Second, the best model from the last scenario will be experienced with an FS-Embedded and one-hot scheme to show the effect of each scheme on the performance of the model.

\begin{table}[htbp]
\caption{Scenario \#1- Comparison between the best model of baseline and transformers in terms of the F1 measure}
\label{table:classes}
\begin{center}
\begin{tabular}{ |c|c|c|c|c|c|c|c| }

\hline
\multicolumn{2}{|c|}{Transformers}&\multicolumn{3}{|c|}{One-Hot}&\multicolumn{3}{|c|}{FS-Embedding}\\

\hline
\multicolumn{1}{|c|}{Class}&\multicolumn{1}{|c|}{Aug}&\multicolumn{1}{|c|}{Precision}&\multicolumn{1}{|c|}{Recall}&\multicolumn{1}{|c|}{F1}&\multicolumn{1}{|c|}{Precision}&\multicolumn{1}{|c|}{Recall}&\multicolumn{1}{|c|}{F1}\\

\hline
0 & - & .94 & .95 & \textbf{.94} & 0.93  &    0.92  &    0.93 \\
\hline
1 & - &.90 & .94 & .92 & 0.91    &  0.93    &  0.92\\
\hline
2 & - &.99 & .99 & .99 & 1.00   &   1.00   & \textbf{1.00}\\
\hline
3 & - &.98 & .97 & \textbf{.98} & 0.91   &   0.90   &   0.90  \\
\hline
4 & - &.92 & .91 & \textbf{.92} & 0.93   &   0.66   &   0.77 \\
\hline
5 & - &.97 & .96 & \textbf{.96} & .95   &   .92   &   .93\\
\hline
6 & - &.95 & .88 & \textbf{.91} & .95   &   .87   &   .91\\
\hline
7 & - & .98 & .99 & .98 & 1.00   &   1.00   &   \textbf{1.00}\\
\hline
8 & - &.94 & .98 & .96 & .94   &   .98   &   .96\\
\hline
9 & - &.99 & .99 & .99 & 1.00   &   1.00   &  \textbf{1.00}\\
\hline
10 & x &.82 & .57 & .67 & .73   &   .96  &   \textbf{.83}\\
\hline
11 &x & .96 & .65 & .78 & .88   &  .91   &  \textbf{.89}\\
\hline
12 &- & .87 & .76 & .81 & .92   &   .80   &   \textbf{.86}\\
\hline
13 &x & .81 & .66 & .73 & .73   &   .94   &   \textbf{.82}\\
\hline
14 &- & .98 & .98 & .98 & .98   &   .98   &   .98\\
\hline
15 &x & .84 & .67 & .74 & .81   &   .87   & \textbf{.84}\\
\hline
16 &- & .90 & .75 & .82 &.88    &  .80    & \textbf{.84}\\
\hline
17 &x & .77 & .87 & .82 & .77  &    .99    &  \textbf{.87}\\
\hline
18 &x & .92 & .67 & .78& .93    &  .94   &   \textbf{.94}\\
\hline
\end{tabular}
\end{center}
\end{table}

\section{evaluation}
This section presents two scenarios to evaluate the performance of our augmentation and FS-Embedding schemes. In the first scenario, we conduct an empirical study of our augmentation method, comparing its effectiveness across three different datasets. The second scenario compares our FS-Embedding scheme to the traditional one-hot encoding method, using the best augmentation strategy from the first scenario.

To fully discover the advantages of our method three sets of datasets are prepared:
\begin{itemize}
    \item Actual data: The exact dataset from section  \ref{datasetsec}
    \item Sampled data: Dataset of section \ref{datasetsec} over-sampled using \cite{Sampling}
    \item Augmented data: Dataset of section \ref{datasetsec} augmented using our method
\end{itemize}
The sampling method described in \cite{Sampling} is a simple yet effective approach for addressing the issue of imbalanced classification. It has been widely utilized in numerous studies, such as in\cite{sharifia}.

The classes NTP, Facebook, Twitter, WindowsUpdate, Instagram, PlayStore, and YouTube were selected for augmentation and over-sampling, as the CRNN network performs the worst on these classes. Additionally, these classes have a low number of samples in the dataset.

The evaluation metrics that are chosen to measure the performance of our approach are those that are mostly used for imbalanced datasets and give an appropriate analysis of the methods that are employed. These metrics are precision, recall, accuracy, and $\mathrm{F_1}$ whose formulas are given in the following.

\begin{equation}
    Precision = \frac{\text{TP}}{\text{TP}+\text{FP}},
\end{equation}

\begin{equation}
    Recall = \frac{\text{TP}}{\text{TP}+\text{FN}}.
\end{equation}
\begin{equation}
    Accuracy = \frac{\text{TP}+\text{TN}}{\text{TP}+\text{TN}+\text{FP}+\text{FN}}.
\end{equation}

The TP, FP, TN, and FN in the above formulas depict true positive, false positive, true negative, and false negative values, respectively.

\begin{equation}
    F_{1}=2*\frac{Precision \times Recall}{Precision + Recall}.
\end{equation}
The $F_1$ measure shows the overall performance of the algorithm on both precision and recall.

A comparison between augmentation schemes on baseline and transformers models is presented in Tables \ref{table:sc1-aug-base}, \ref{table:sc1-aug-trans}. It can be seen that our augmentation scheme outperforms in all classes compared with other schemes.

\begin{table}[htbp]
\caption{Scenario \#1- Comparison between different data augmentation schemes over baseline model in terms of the F1 measure}
\label{table:sc1-aug-base}
\begin{center}
\begin{tabular}{ |c|c|c|c|c| }

\hline
\multicolumn{1}{|c|}{Baseline}&\multicolumn{2}{|c|}{Augmented}&\multicolumn{1}{|c|}{Sampled}&\multicolumn{1}{|c|}{Actual}\\

\hline
\multicolumn{1}{|c|}{Class}&\multicolumn{1}{|c|}{Aug}&\multicolumn{1}{|c|}{F1}&\multicolumn{1}{|c|}{F1}&\multicolumn{1}{|c|}{F1}\\

\hline
0 & - & \textbf{0.94} & 0.91 & 0.90 \\
\hline
1 & - & \textbf{0.92} & 0.87 & 0.79\\
\hline
2 & - &\textbf{0.99} & 0.98 & 0.50\\
\hline
3 & - &\textbf{0.98} & 0.95 & 0.97\\
\hline
4 & - &\textbf{0.92} & 0.87 & 0.88\\
\hline
5 & - &\textbf{0.96} & 0.91 & 0.94\\
\hline
6 & - &\textbf{0.91} & 0.86 & 0.88\\
\hline
7 & - & \textbf{0.98} & 0.96 & 0.97\\
\hline
8 & - &\textbf{0.96} & 0.95 & 0.94\\
\hline
9 & - &0.99 & 0.99 & 0.98\\
\hline
10 & x &\textbf{0.67} & 0.54 & 0.59\\
\hline
11 &x &\textbf{0.78}  & 0.64 & 0.72\\
\hline
12 &- &\textbf{0.81}  & 0.71 & 0.68\\
\hline
13 &x &\textbf{0.73}  & 0.52 & 0.47\\
\hline
14 &- &\textbf{0.98}  & 0.97 & 0.91\\
\hline
15 &x &\textbf{0.74}  & 0.55 & 0.58\\
\hline
16 &- & \textbf{0.82} & 0.75 & 0.76\\
\hline
17 &x & \textbf{0.82} & 0.30 & 0.24\\
\hline
18 &x & \textbf{0.78} & 0.69 & 0.72\\
\hline
\end{tabular}
\end{center}
\end{table}

\begin{table}[htbp]
\caption{Scenario \#1- Comparison between different data augmentation schemes over transformers in terms of the F1 measure}
\label{table:sc1-aug-trans}
\begin{center}
\begin{tabular}{ |c|c|c|c|c| }

\hline
\multicolumn{1}{|c|}{Transformers}&\multicolumn{2}{|c|}{Augmented}&\multicolumn{1}{|c|}{Sampled}&\multicolumn{1}{|c|}{Actual}\\

\hline
\multicolumn{1}{|c|}{Class}&\multicolumn{1}{|c|}{Aug}&\multicolumn{1}{|c|}{F1}&\multicolumn{1}{|c|}{F1}&\multicolumn{1}{|c|}{F1}\\

\hline
0 & - & \textbf{0.93} & 0.91 & 0.89 \\
\hline
1 & - &\textbf{0.92} & 0.87 & 0.80\\
\hline
2 & - &\textbf{1.00} & 0.99 & 0.56\\
\hline
3 & - &\textbf{0.98} & 0.95 & 0.97\\
\hline
4 & - &\textbf{0.90} & 0.88 & 0.88\\
\hline
5 & - &\textbf{0.96} & 0.92 & 0.98\\
\hline
6 & - &\textbf{0.89} & 0.86 & 0.86\\
\hline
7 & - & \textbf{0.99} & 0.98 & 0.97\\
\hline
8 & - &\textbf{0.96} & 0.92 & 0.90\\
\hline
9 & - &0.98 & 0.98 & 0.96\\
\hline
10 & x &\textbf{0.77} & 0.54 & 0.64\\
\hline
11 &x &\textbf{0.82}  & 0.74 & 0.70\\
\hline
12 &- & \textbf{0.83} & 0.34 & 0.70\\
\hline
13 &x &\textbf{0.73} & 0.72 & 0.54\\
\hline
14 &- & \textbf{0.98} & 0.84 & 0.83\\
\hline
15 &x & \textbf{0.88} & 0.71 & 0.63\\
\hline
16 &- &\textbf{0.81} & 0.78 & 0.74\\
\hline
17 &x &\textbf{0.80}  & 0.45 & 0.21\\
\hline
18 &x &\textbf{0.89}& 0.77 & 0.79\\
\hline
\end{tabular}
\end{center}
\end{table}

Table \ref{table:sc1-aug-base-trans} compares the performance of transformers with the baseline model across various augmentation schemes, focusing on the number of parameters, epochs, and accuracy measurements. This comparison shows that the transformer model improves both the speed of convergence and accuracy compared to the baseline model. Table \ref{table:sc1-aug-base-trans-details} further compares the performance of the best models, which are selected from the augmentation schemes of both the baseline and transformer models, in terms of precision, recall, and $F_1$. As we can see, the transformer model increases $F_1$ in augmented data while also improving overall performance in other evaluation metrics. Therefore, we proceed with using the transformer model and our augmentation scheme to evaluate the FS-Embedding scheme in the second scenario.

\begin{table}[htbp]
\caption{Scenario \#1- Comparison between augmentation, sampling, and no scheme over model accuracy, convergence speed,d and number of parameters}
\label{table:sc1-aug-base-trans}
\begin{center}
\begin{tabular}{ |c|c|c|c|c| }

\hline
\multicolumn{1}{|c|}{Model}&\multicolumn{1}{|c|}{Scheme}&\multicolumn{1}{|c|}{\#Parameters}&\multicolumn{1}{|c|}{\#Epochs}&\multicolumn{1}{|c|}{Acc.}\\

\hline
Baseline & - & $\approx$ 491K & 1000 & 87.33\\

\hline
Baseline & Sampled & $\approx$ 491K & 1000 & 90.54\\

\hline
Baseline & Augmented & $\approx$ 491K & 1000 & 93.89\\

\hline
Transformers & - & $\approx$ 458K & 300 & 89.61\\

\hline
Transformers & Sampled & $\approx$ 458K & 300 & 90.62\\

\hline
Transformers & Augmented & $\approx$ 458K & 300 & \textbf{94.96}\\

\hline

\end{tabular}
\end{center}
\end{table}

\begin{table}[htbp]
\caption{Scenario \#1- Comparison between the best model of baseline and transformers in terms of the precision, recall, and F1 measure}
\label{table:sc1-aug-base-trans-details}
\begin{center}
\begin{tabular}{ |c|c|c|c|c|c|c|c| }

\hline
\multicolumn{2}{|c|}{}&\multicolumn{3}{|c|}{Baseline}&\multicolumn{3}{|c|}{Transformers}\\

\hline
\multicolumn{1}{|c|}{Class}&\multicolumn{1}{|c|}{Aug}&\multicolumn{1}{|c|}{Precision}&\multicolumn{1}{|c|}{Recall}&\multicolumn{1}{|c|}{F1}&\multicolumn{1}{|c|}{Precision}&\multicolumn{1}{|c|}{Recall}&\multicolumn{1}{|c|}{F1}\\

\hline
0 & - & 0.94 & 0.95 & \textbf{0.94} & 0.93 & 0.92 & 0.93 \\
\hline
1 & - &0.90 & 0.94 & 0.92 & 0.91 & 0.93 & 0.92\\
\hline
2 & - &0.99 & 0.99 & 0.99 & 1.00 & 1.00 & \textbf{1.00}\\
\hline
3 & - &0.98 & 0.97 & 0.98 & 0.99 & 0.96   & 0.98\\
\hline
4 & - &0.92 & 0.91 & \textbf{0.92} & 0.91   &   0.90   &   0.90\\
\hline
5 & - &0.97 & 0.96 & 0.96 & 0.98 & 0.95 & 0.96\\
\hline
6 & - &0.95 & 0.88 & \textbf{0.91} & 0.98 & 0.81 & 0.89\\
\hline
7 & - & 0.98 & 0.99 & 0.98 & 1.00 & 0.98 & \textbf{0.99} \\
\hline
8 & - &0.94 & 0.98 & 0.96 & 0.95   &   0.98  & \textbf{0.96}\\
\hline
9 & - &0.99 & 0.99 & \textbf{0.99} & 0.99& 0.97 & 0.98\\
\hline
10 & x &0.82 & 0.57 & 0.67 & 0.67 & 0.92 & \textbf{0.77}\\
\hline
11 &x & 0.96 & 0.65 & 0.78 & 0.78 & 0.85 & \textbf{0.82}\\
\hline
12 &- & 0.87 & 0.76 & 0.81 & 0.91 & 0.75 & \textbf{0.83}\\
\hline
13 &x & 0.81 & 0.66 & 0.73 & 0.60 & 0.94 & 0.73 \\
\hline
14 &- & 0.98 & 0.98 & 0.98 & 0.98 & 0.99 & \textbf{0.98}\\
\hline
15 & x & 0.84 & 0.67 & 0.74 & 0.89 & 0.87 & \textbf{0.88}\\
\hline
16 &- & 0.90 & 0.75 & \textbf{0.82} & 0.83  &    0.80   &   0.81\\
\hline
17 &x & 0.77 & 0.87 & \textbf{0.82} & 0.69   &   0.95   &   0.80\\
\hline
18 &x & 0.92 & 0.67 & 0.78& 0.87 & 0.90 & \textbf{0.89}\\
\hline
\end{tabular}
\end{center}
\end{table}

The second scenario investigates the effect of FS-Embedding on the transformer model. In this scenario, we first train the model using the one-hot scheme to enable further comparison. Table \ref{table:trans-one-hot} shows the models trained with various parameters less than 300 epochs. An accuracy of 94.96\% is achieved with nearly 458K parameters, while FS-Embedding uses less than 50\% of the parameters to achieve higher accuracy. Table \ref{table:one-hot-fs-trans} provides more details comparing the one-hot and FS-Embedding schemes on the transformer model.

\begin{table}[htbp]
\caption{Scenario \#2- Comparison between transformers models with various parameters under the one-hot scheme in terms of accuracy}
\label{table:trans-one-hot}
\begin{center}
\begin{tabular}{ |c|c|c|c|c| }
\hline
\multicolumn{5}{|c|}{Transformers}\\

\hline
\multicolumn{1}{|c|}{Augmentation}&\multicolumn{1}{|c|}{Embedding}&\multicolumn{1}{|c|}{\#Parameters}&\multicolumn{1}{|c|}{\#Epochs}&\multicolumn{1}{|c|}{Acc.}\\

\hline
Yes & One-Hot & $\approx$ 33K & 300 & 86.12\\

\hline
Yes & One-Hot & $\approx$ 77K & 300 & 89.43\\

\hline
Yes & One-Hot & $\approx$ 213K & 300 & 90.21\\

\hline
Yes & One-Hot & $\approx$ 458K & 300 & 94.96\\

\hline

\end{tabular}
\end{center}
\end{table}

\begin{table}[htbp]
\caption{Scenario \#2 – Comparison of transformer models with various parameters under the one-hot scheme in terms of accuracy.}
\label{table:one-hot-fs-trans}
\begin{center}
\begin{tabular}{ |c|c|c|c|c| }
\hline
\multicolumn{5}{|c|}{Transformers}\\

\hline
\multicolumn{1}{|c|}{Augmentation}&\multicolumn{1}{|c|}{Embedding}&\multicolumn{1}{|c|}{\#Parameters}&\multicolumn{1}{|c|}{\#Epochs}&\multicolumn{1}{|c|}{Acc.}\\

\hline
Yes & One-Hot & $\approx$ 458K & 300 & 94.96\\

\hline
Yes & FS & $\approx$ 35K & 300 & 94.15\\

\hline
Yes & FS & $\approx$ 77K & 300 & 95.14\\

\hline
Yes & FS & $\approx$ 213K & 300 & 95.63\\

\hline

\end{tabular}
\end{center}
\end{table}

\section{conclusion}
First, we proposed an augmentation scheme to address imbalanced network traffic classification using real traffic traces, based on LSTM and KDE. Next, we introduced a novel embedding method based on the transformer encoder layer to extract valuable features from ports and packet directions in the flow. We then tested our augmentation and embedding pipeline on a state-of-the-art deep-learning model. Our experiments demonstrate that the proposed pipeline enhances the accuracy and generalization of our data. Additionally, our embedding scheme outperforms the one-hot method by reducing the network’s complexity.

\bibliographystyle{unsrt}
\bibliography{references}

\end{document}